\documentclass[a4paper,twoside]{article}
\usepackage{booktabs}
\usepackage{epsfig}
\usepackage{subcaption}
\usepackage{calc}
\usepackage{amssymb}
\usepackage{amstext}
\usepackage{amsmath,bm}

\newcommand{\wi}[1]{\boldsymbol{w}_{#1}}   
\newcommand{\wrnd}[1]{\boldsymbol{w}^{(#1)}} 
\usepackage{amsthm}
\usepackage{multicol}
\usepackage{pslatex}
\usepackage[hidelinks]{hyperref}
\usepackage{verbatim}
\usepackage{multirow}
\usepackage[bottom]{footmisc}
\usepackage[margin=0.5in]{geometry}
\usepackage{SCITEPRESS}     
\usepackage{natbib}
\usepackage{balance}
\usepackage{acronym}
\usepackage{lscape}
\usepackage{graphicx} 
\usepackage{caption} 
\usepackage{tabularx}
\begin{document}
\acrodef{AI}{Artificial Intelligence}

\acrodef{FL}{Federated Learning}

\acrodef{PSO}{Particle Swarm Optimization}

\acrodef{CNN}{Convolutional Neural Network}

\acrodef{CNNs}{Convolutional Neural Networks}

\acrodef{CT}{Computed Tomography}

\acrodef{non-IID}{non-independent and identically distributed}

\acrodef{Biglycan}{Breast Cancer Histopathological Dataset}

\acrodef{GANs}{Generative Adversarial Networks}

\acrodef{LC25000}{Lung and Colon Cancer Histopathological Images}

\acrodef{IHC}{Immunohistochemistry}

\acrodef{MRI}{Magnetic Resonance Imaging}

\acrodef{ResNet}{Residual Network}

\acrodef{AUC}{Area Under the Curve}

\acrodef{FedAvg}{Federated Averaging}

\acrodef{TTA}{Test Time Augmentation}

\title{Exploiting Test-Time Augmentation in Federated Learning for Brain Tumor MRI Classification}


\author{\authorname{Thamara Leandra de Deus Melo\sup{1}, Rodrigo Moreira\sup{1}\orcidAuthor{0000-0002-9328-8618}, \\Larissa Ferreira {Rodrigues Moreira}\sup{1}\orcidAuthor{0000-0001-8947-9182}, André R. Backes\sup{2}\orcidAuthor{0000-0002-7486-4253}}
\affiliation{\sup{1}Institute of Exact and Technological Sciences, Federal University of Viçosa - UFV, Rio Paranaíba-MG, Brazil}
\affiliation{\sup{2}Department of Computing, Federal University of São Carlos, São Carlos-SP, Brazil}
\email{\{thamara.melo, rodrigo, larissa.f.rodrigues\}@ufv.br, arbackes@yahoo.com.br}
}

\keywords{Brain tumors, Federated Learning, Test-Time Augmentation, Image classification}

\abstract{
Efficient brain tumor diagnosis is crucial for early treatment; however, it is challenging because of lesion variability and image complexity. We evaluated convolutional neural networks (CNNs) in a federated learning (FL) setting, comparing models trained on original versus preprocessed MRI images (resizing, grayscale conversion, normalization, filtering, and histogram equalization). Preprocessing alone yielded negligible gains; combined with test-time augmentation (TTA), it delivered consistent, statistically significant improvements in federated MRI classification ($p<0.001$). In practice, TTA should be the default inference strategy in FL-based medical imaging; when the computational budget permits, pairing TTA with light preprocessing provides additional reliable gains. 
}

\onecolumn \maketitle \normalsize \setcounter{footnote}{0} \vfill

\acrodef{AI}{Artificial Intelligence}

\acrodef{FL}{Federated Learning}

\acrodef{PSO}{Particle Swarm Optimization}

\acrodef{CNN}{Convolutional Neural Network}

\acrodef{CNNs}{Convolutional Neural Networks}

\acrodef{CT}{Computed Tomography}

\acrodef{non-IID}{non-independent and identically distributed}

\acrodef{Biglycan}{Breast Cancer Histopathological Dataset}

\acrodef{GANs}{Generative Adversarial Networks}

\acrodef{LC25000}{Lung and Colon Cancer Histopathological Images}

\acrodef{IHC}{Immunohistochemistry}

\acrodef{MRI}{Magnetic Resonance Imaging}

\acrodef{ResNet}{Residual Network}

\acrodef{AUC}{Area Under the Curve}

\acrodef{FedAvg}{Federated Averaging}

\acrodef{TTA}{Test Time Augmentation}

\section{\uppercase{Introduction}}

Brain tumors are among the most aggressive neurological disorders with high morbidity and mortality. Diagnosis is difficult because lesions exhibit heterogeneous morphologies, variable intensity profiles, and complex appearances on \ac{MRI} \citep{stupp2005,louis2021,Carmo2025}. Although \ac{MRI} is the gold standard for noninvasive assessment, interpretation depends on specialist expertise, which introduces variability and limits scalability in routine clinical practice \citep{wen2020}.

Recent advances in \ac{AI}, particularly in \ac{CNN}-based models, have improved automated tumor assessment by learning discriminative features directly from images \citep{Goodfellow2016, Rodrigues2017, RodriguesMoreira2025}. However, privacy regulations and institutional policies often preclude centralizing of patient data across hospitals. \ac{FL} addresses this barrier by training shared models locally at each site without exposing raw data, promoting generalization across heterogeneous cohorts while preserving confidentiality \citep{mcmahan2017,konecn2016, Leonardo2025}.

Despite the growing use of \ac{FL} in medical imaging, the impact of common preprocessing pipelines on federated performance for brain tumor \ac{MRI} classification remains unclear. Standard operations (e.g., resizing, grayscale conversion, normalization, filtering, and histogram equalization) may harmonize inputs across clients; however, they can also suppress subtle, clinically relevant cues. At the same time, the potential of \ac{TTA} at inference, known to reduce prediction variance in centralized settings, has not been examined under federated conditions, where client heterogeneity (scanners, protocols, and demographics) is the norm \citep{Wang2019,Islam2024}.

\ac{TTA} is model-agnostic, requires no retraining, and can be deployed unilaterally at inference time by each client, all of which are attractive properties in \ac{FL} where retraining is costly, communication budgets are tight, and data cannot be pooled. If effective, \ac{TTA} offers a low-overhead mechanism to improve robustness and stabilize predictions against interclient distribution shifts and small-sample effects without modifying the federated optimization process.

Guided by the gaps identified, we raise the following research questions (RQs) to structure our investigation:

\begin{itemize}
    \item \textbf{RQ1:} Does preprocessing help or hinder brain-tumor classification in \ac{FL} when compared with training on original (minimally processed) \ac{MRI} images?
    \item \textbf{RQ2:} To what extent does \ac{TTA} improve prediction robustness and accuracy during federated inference across heterogeneous clients?
\end{itemize}

This study addresses these gaps by proposing an \ac{FL} framework for brain tumor \ac{MRI} classification that compares original and preprocessed images and integrates \ac{TTA} during inference. Our contributions are as follows: (i) we evaluate the influence of preprocessing pipelines on federated performance, (ii) we exploit \ac{TTA} to enhance prediction stability and accuracy, and (iii) quantify effects via statistical hypothesis testing.  

Our results show that preserving native \ac{MRI} characteristics yields better federated performance than aggressively preprocessed inputs and that \ac{TTA} provides a consistent, low-cost robustness boost. These findings provide practical data-retentive design choices for decentralized medical \ac{AI} systems.

The remainder of this paper is organized as follows. Section \ref{sec:rw} reviews related work. Section \ref{sec:method} describes the proposed method, including the \ac{MRI} dataset, preprocessing pipeline, federated setup, and integration of \ac{TTA}. Section \ref{sec:results} presents and discuss the experimental results. Finally, Section \ref{sec:conclusion} concludes the paper and outlines future research directions.

\section{\uppercase{Related Work}}\label{sec:rw}

The detection and classification of brain tumors in \ac{MRI} images has been the subject of extensive investigation, given the severity of this condition and the need for early and accurate diagnoses. Machine learning and deep learning techniques have been prominently applied in this context, especially \ac{CNNs}, which have proven efficient in extracting discriminative patterns in medical images \cite{Solanki2023, Siddique2021}.  

\cite{Shah2022} presented a model based on the EfficientNet-B0 architecture for brain tumor classification, combined with \textit{fine-tuning} techniques and data augmentation. The authors compared the model with traditional architectures such as VGG16, InceptionV3, Xception, ResNet50, and InceptionResNetV2. The proposed model outperformed all baselines, achieving an accuracy of 98.87\% and an AUC of 0.988.  

\cite{khaliki2024} compared the performance of conventional CNNs and pretrained models such as VGG19, EfficientNetB4, and InceptionV3 on Kaggle dataset images. The results demonstrated that the use of \textit{transfer learning} significantly improves classification, particularly in scenarios with limited datasets.  

\cite{kaur2025} explored the use of ResNet152 and GoogleNet for feature extraction, combined with traditional classifiers such as SVM, KNN, CART, and GNB. The pipeline also included preprocessing, PCA, and data augmentation. The study achieved an accuracy of 98.53\% and an F1-score of 97.4\%, highlighting the relevance of image preparation and architectural choice for consistent results.  

In the field of \ac{FL}, \cite{zhou2024distributed} demonstrated the feasibility of decentralized training across multiple medical institutions using the BRATS dataset. The study evaluated several CNNs, including ResNet50, VGG16, EfficientNet, and AlexNet, achieving accuracy above 95\%. Similarly, \cite{albalawi2024} integrated \ac{FL} and \textit{transfer learning} with a modified VGG16, obtaining 98\% accuracy and an F1-score above 0.95 while preserving the privacy of sensitive data.  

Thus, the literature highlights significant advances in brain tumor classification in both centralized and distributed scenarios. Unlike the aforementioned works, this study focuses on comparing original and preprocessed images in a \ac{FL} environment, using ResNet18 to evaluate the impact of preprocessing on classification performance.  

Table~\ref{tab:trabalhosrelacionados} summarizes the main characteristics of the mentioned studies on brain tumor MRI classification. To improve readability, the table uses a checklist notation: \checkmark indicates that the feature or technique was explicitly employed in the study, whereas $\times$ indicates its absence. This compact format allows direct comparison of datasets, model backbones, preprocessing, FL usage, and evaluation metrics.

\begin{table*}[ht]
\centering
\caption{Comparison with related work.}
\renewcommand{\arraystretch}{1.2}
\label{tab:trabalhosrelacionados}
\resizebox{\textwidth}{!}{
\renewcommand{\arraystretch}{1.2}
\begin{tabular}{lp{3cm}p{3cm}ccc}
\toprule 
\textbf{Work} & \textbf{Dataset} & \textbf{Method} & \textbf{Pre/Aug.} & \textbf{FL} & \textbf{Metrics}\\
\midrule
\cite{Shah2022}            & BRATS 2015     & EfficientNet-B0 & \checkmark & $\times$ & Acc, AUC \\ \hline
\cite{Jiang2022}           & BRATS 2019–21  & 3D Swin Transformer     & \checkmark & $\times$ & Dice, HD \\ \hline
\cite{khaliki2024}         & Kaggle (MRI)   & VGG19, EfficientNetB4, InceptionV3  & \checkmark & $\times$ & Acc, F1, AUC \\ \hline
\cite{Shaikh2025}          & BRATS 2020–23  & Ensemble of CNNs  & \checkmark & $\times$ & Acc, F1, Dice \\ \hline
\cite{kaur2025}            & Br35 (Kaggle)  & ResNet152, GoogleNet    & \checkmark & $\times$ & Acc, F1 \\ \hline
\cite{zhou2024distributed} & BRATS 2015–18  & ResNet50, VGG, EffNet   & $\times$ & \checkmark & Acc, F1, AUC \\ \hline
\cite{albalawi2024}        & Figshare, Br35H& VGG16 + FL (10 clients) & \checkmark & \checkmark & Acc, F1 \\ \hline
\textbf{Ours}              & \textbf{Kaggle MRI}     & \textbf{ResNet18 + TTA }         & \checkmark & \checkmark & \textbf{Acc, F1, Recall, Precision} \\
\bottomrule
\end{tabular}
}
\end{table*}

Unlike previous studies, this study advances the field by comparing original and preprocessed \ac{MRI} images in a federated setting, revealing that preserving the raw image structure can yield superior performance. We are among the first to exploit \ac{TTA} within \ac{FL} for brain-tumor classification, demonstrating its role in improving prediction stability and highlighting the differences between data preparation strategies. We reinforce our findings with rigorous statistical hypothesis testing, providing quantitative evidence of effect size and significance, which is often overlooked in related studies. 

\section{\uppercase{Material and Methods}}\label{sec:method}

\begin{figure}[ht]
  \centering
  \includegraphics[width=1\columnwidth]{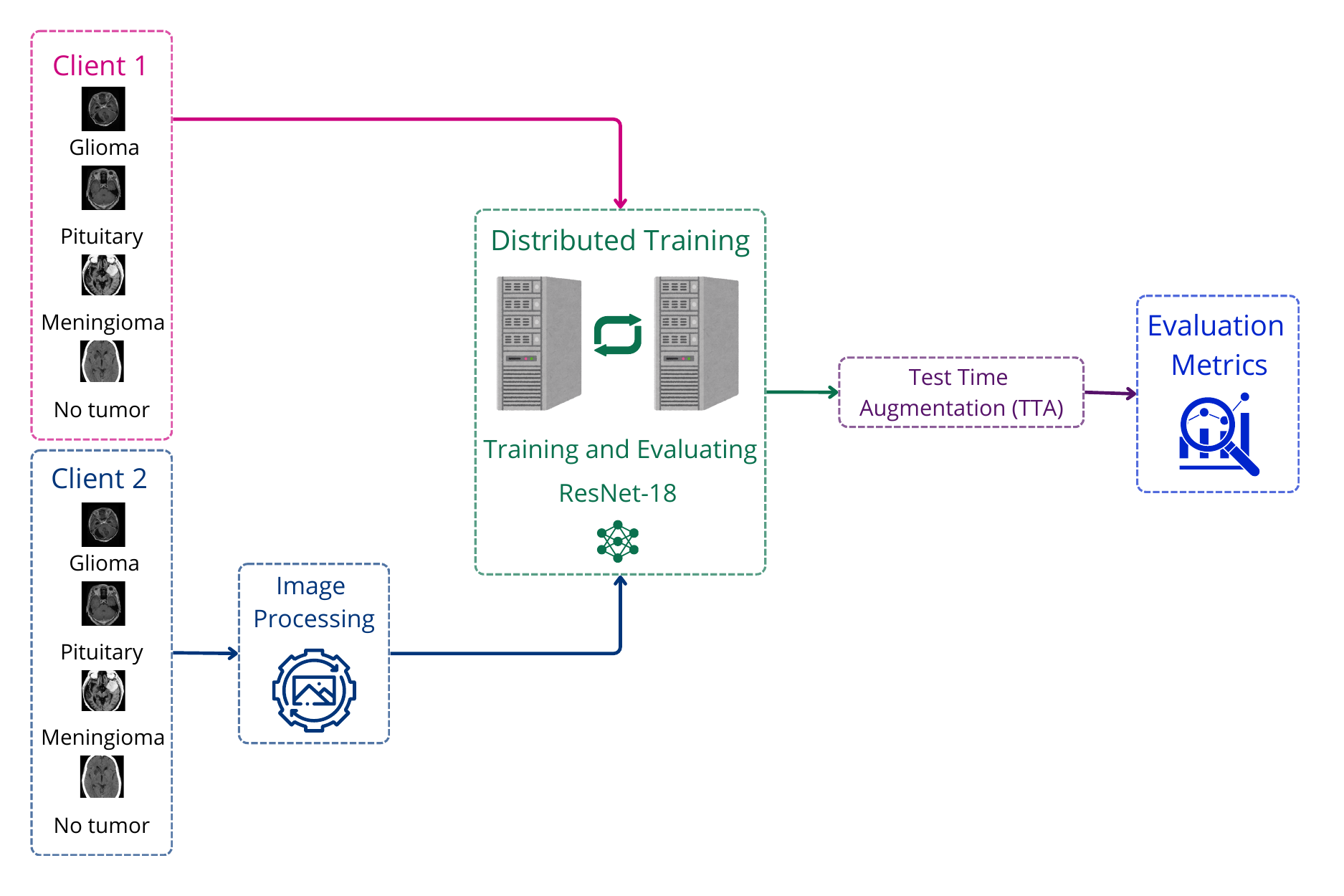}
  \caption{Proposed Method.}
  \label{fig:proposed-method}
\end{figure}

We investigated how input preprocessing and test-time augmentation (TTA) affect brain–tumor MRI classification under federated learning (FL). Figure~\ref{fig:proposed-method} shows an end-to-end workflow. 

Two clients participated in FL: Client~1 received the original (minimally processed) images, whereas Client~2 consumed the preprocessed images produced by a standard preprocessing pipeline. Each client trains a local ResNet-18 on its private subset; after each communication round $r$, local parameters $\wi{i}^{(r)}$ are sent to the server, which aggregates them via FedAvg to obtain the global model $\wrnd{r+1}$. 

Once training converges, inference is performed with TTA. For each test image, $K$ stochastic views are generated, scored by the global model, and combined (e.g., by averaging) into a final prediction. 

\subsection{Image Dataset}

We use the public Brain \ac{MRI} dataset from Kaggle\footnote{\url{https://www.kaggle.com/datasets/masoudnickparvar/brain-tumor-mri-dataset}}, a widely adopted in the literature for brain-tumor classification. The dataset consists of 2D axial brain \ac{MRI} slices, provided as raster images with variable spatial resolutions and aspect ratios. 

The images were labeled into four categories:
no tumor, glioma, meningioma, and pituitary. In the original release, the class counts are
no tumor (2000), glioma (1621), meningioma (1645), and pituitary (1757), totaling 7023 images. Table \ref{tab:exemplos_classes} presents the distribution of instances per class along with representative visual examples.  

\begin{table}[ht]
\centering
\caption{Class distribution with visual examples.}
\renewcommand{\arraystretch}{1.5}
\label{tab:exemplos_classes}
\begin{tabular}{lcl}
\hline
\textbf{Class} & \textbf{Quantity} & \textbf{Example}                                                 \\ \hline
No Tumor       & 2000              & \includegraphics[width=0.15\columnwidth]{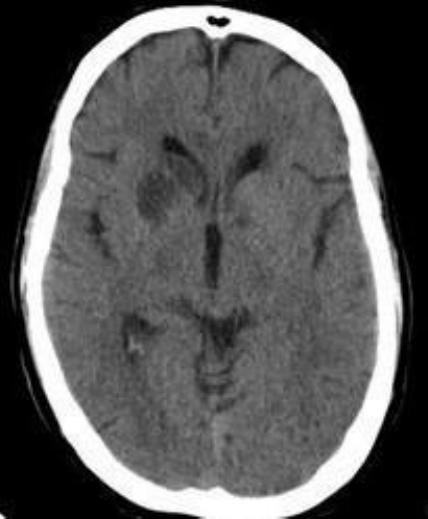}   \\
Glioma         & 1621              & \includegraphics[width=0.15\columnwidth]{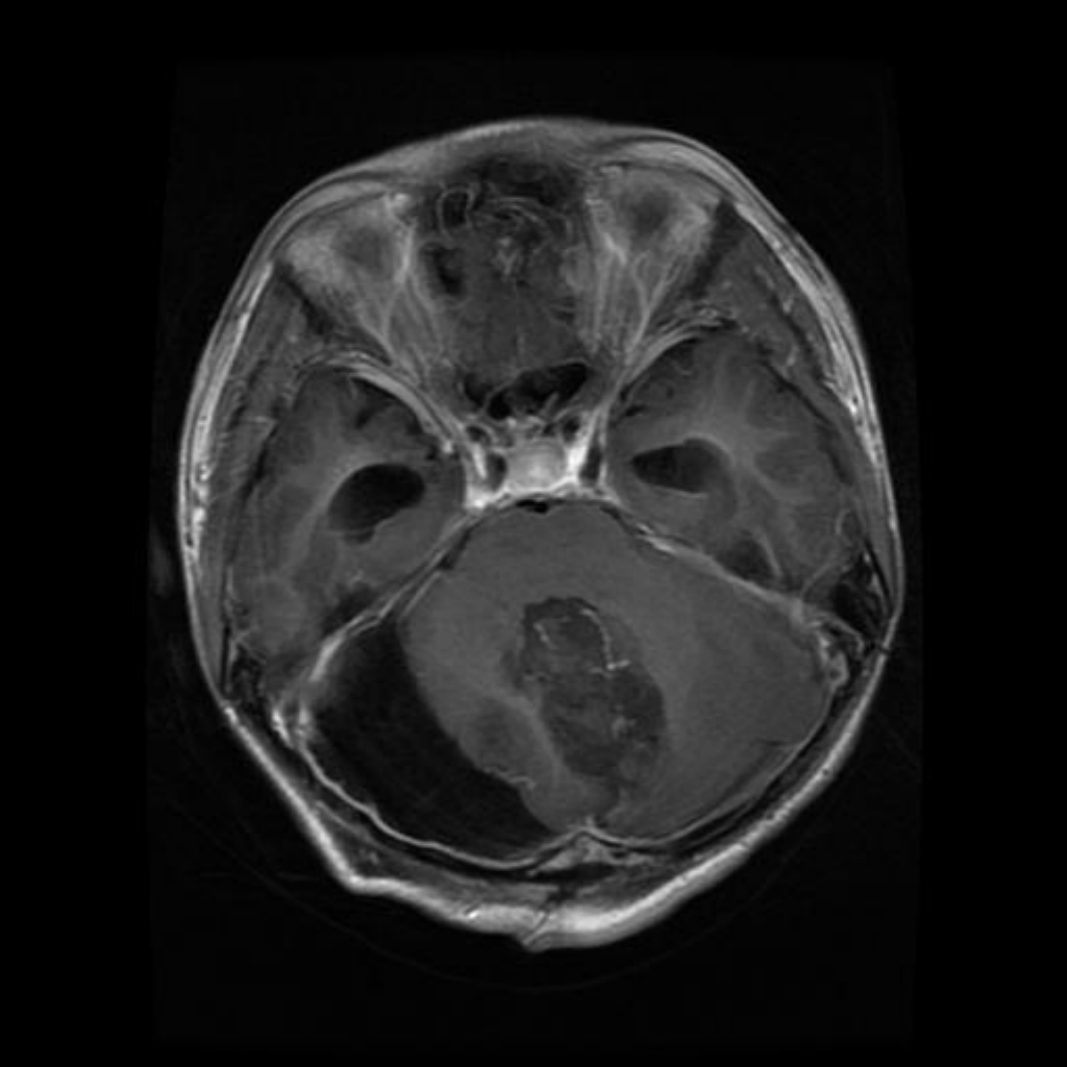}     \\
Meningioma     & 1645              & \includegraphics[width=0.15\columnwidth]{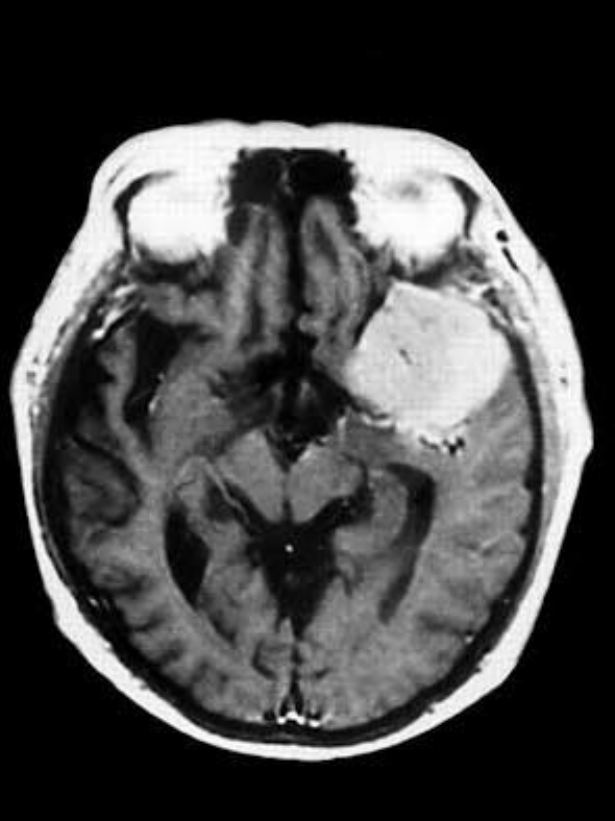} \\
Pituitary      & 1757              & \includegraphics[width=0.15\columnwidth]{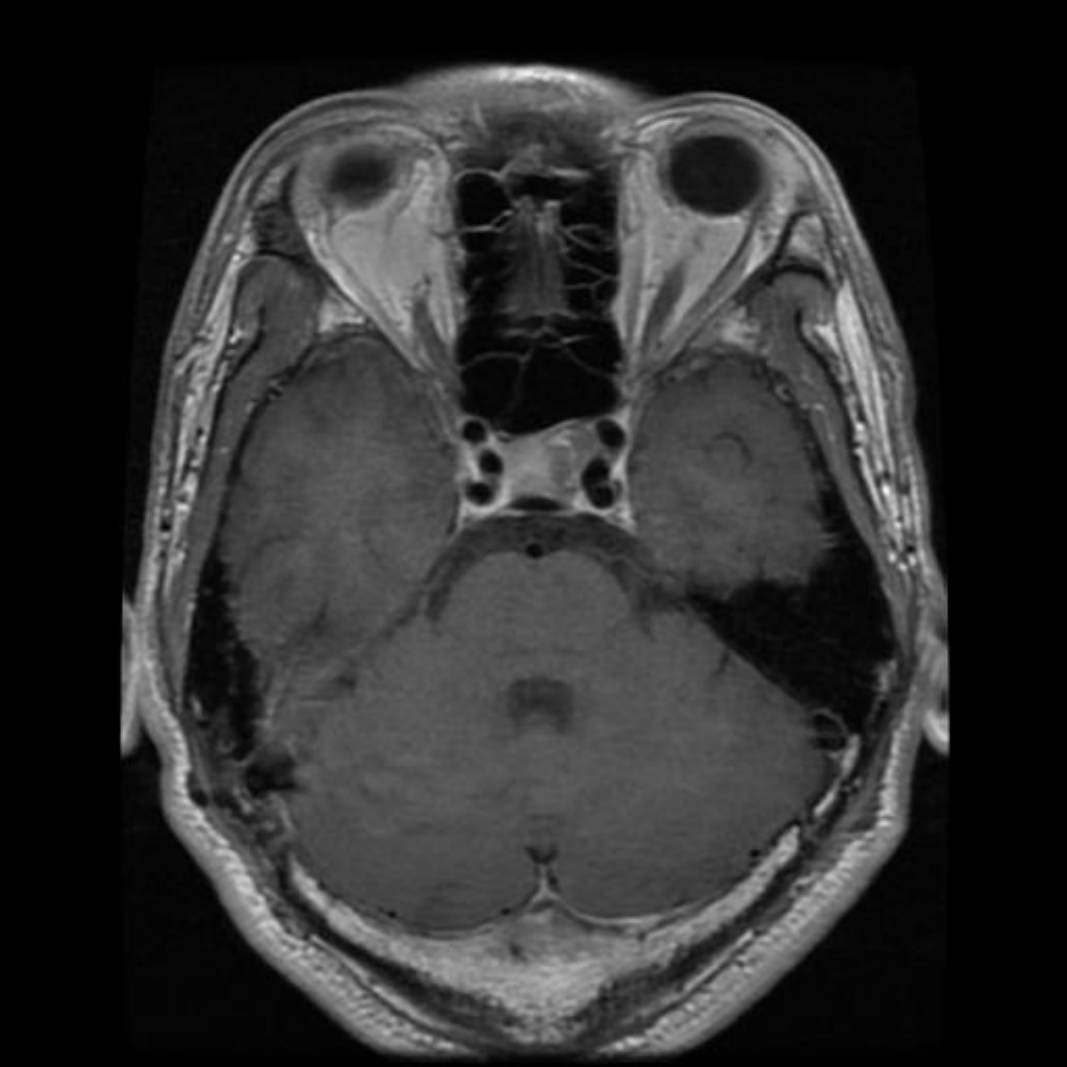} \\ \hline
\textbf{Total} & \textbf{7023}     &                                                                  \\ \hline
\end{tabular}
\end{table}

To mitigate information leakage and prevent optimistic bias, we removed duplicate data prior to splitting the dataset. We identified exact bit-for-bit duplicates by calculating the SHA-256 hash for each file, finding all byte-identical matches regardless of their class. A total of 194 duplicate images were removed, resulting in a final working dataset of 6,829 unique images for all experiments~\citep{shs2012}.

\subsection{ResNet-18 Backbone}

We adopted ResNet-18 as the convolutional backbone for all experiments. ResNet implements residual learning and instead of learning direct mapping $H(x)$, each block learns a residual function $F(x) = H(x) - x$ and reintroduces the input through an identity shortcut $x + F(x)$. This design facilitates optimization in deeper models by preserving the gradient flow and mitigating the vanishing gradients \citep{He2016}.

\begin{figure}[ht]
  \centering
  \includegraphics[width=1\linewidth]{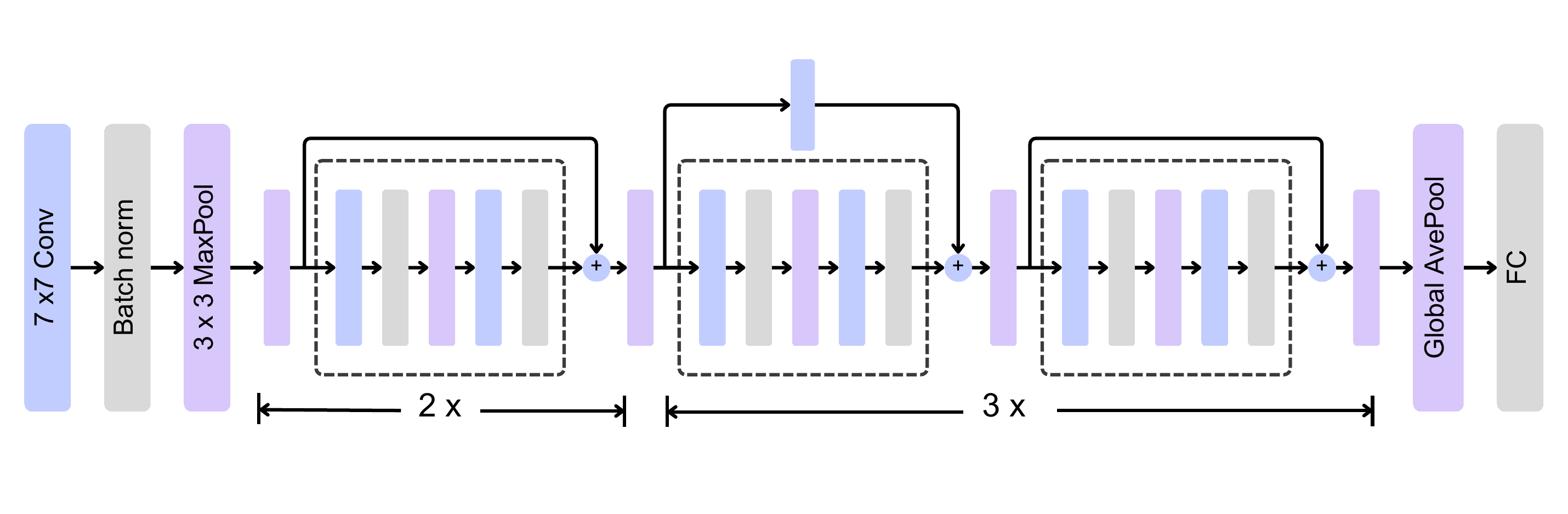}
  \caption{Schematic of the ResNet\,18 architecture with basic residual blocks and four stages of increasing channel width.}
  \label{fig:resnet18}
\end{figure}

ResNet-18 comprises four stages of basic blocks with $3{\times}3$ convolutions, each followed by batch normalization and ReLU. Downsampling was performed using stride\,2 in the first block of each stage. The channel widths per stage were $[64, 128, 256, 512]$ with block counts $[2, 2, 2, 2]$. A $7{\times}7$ convolution and max pooling form the stem, and the network ends with global average pooling and a fully connected classifier (Fig.\ref{fig:resnet18}). 

This \ac{CNN} has approximately 11.7\,M parameters, which offers a favorable balance between the representational capacity and communication cost in \ac{FL} rounds~\citep{RodriguesMoreira2025}. Local updates are computed for each client, and the resulting parameters are aggregated by FedAvg on the server. ResNet-18 is sufficiently compact for fast rounds in \ac{FL}. When the hardware allows, we use 16-bit (FP16) weights, which take approximately half the memory and bandwidth of 32-bit (FP32) to further speed up communication without changing the architecture. This keeps both the original and preprocessed pipelines directly comparable.

\subsection{Federated Learning}

Federated Learning (\ac{FL}) is a decentralized training paradigm in which data never leave the client devices (e.g., hospitals); only model updates are transmitted to a coordinating server that computes a global model \citep{mcmahan2017}. This design aligns with medical privacy requirements by avoiding the centralization of sensitive \ac{MRI} data, while still enabling collaborative learning across institutions \citep{konecn2016, Leonardo2025}.

\begin{figure}[!ht]
  \centering
  \includegraphics[width=1\columnwidth]{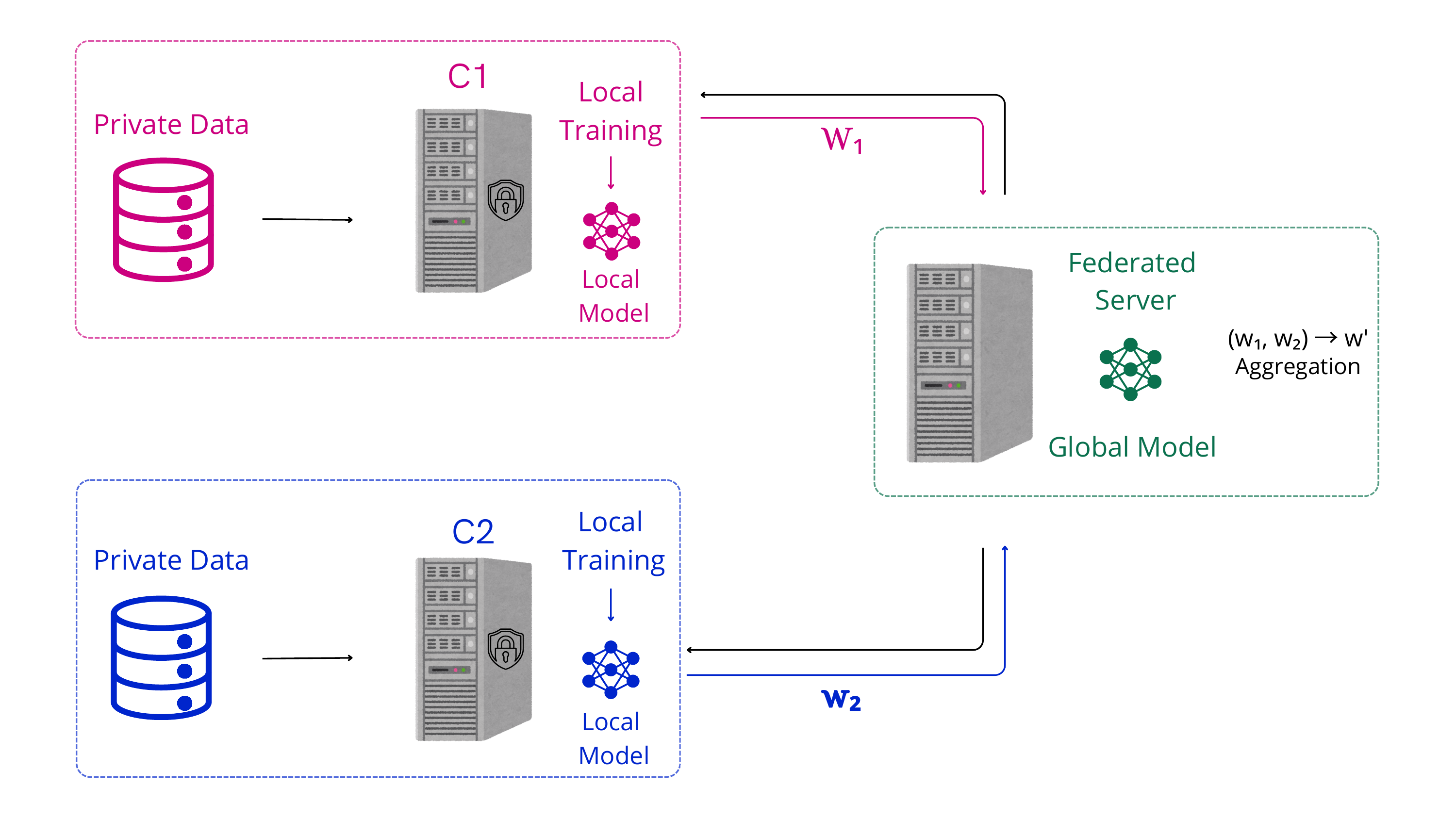}
  \caption{Basic \ac{FL} topology.}
  \label{fig:estrutura-fl}
\end{figure}

Training proceeds in synchronous communication rounds \(r \in \{0,\dots,R-1\}\).
At the start of round \(r\), the server broadcasts the current global parameters \(\mathbf{w}^{(r)}\) to the participating clients \(\mathcal{C}_r\).
Each client \(i \in \mathcal{C}_r\) performs local training and returns \(\mathbf{w}^{(r)}_i\).
The server updates the global model via sample-size–weighted averaging (Equation \ref{eq:fedavg}).
\begin{equation}
\mathbf{w}^{(r+1)}
=
\frac{\displaystyle \sum_{i \in \mathcal{C}_r} n_i\, \mathbf{w}^{(r)}_i}
{\displaystyle \sum_{j \in \mathcal{C}_r} n_j}.
\label{eq:fedavg}
\end{equation}
Where \(n_i\) denotes the number of training samples held by client \(i\), and \(\mathcal{C}_r\) is the set of clients participating in round \(r\).
Figure~\ref{fig:estrutura-fl} depicts the standard \ac{FL} topology with clients and servers.

Medical datasets are typically fragmented across sites, collected using heterogeneous scanners and protocols, and cannot be pooled because of consent and regulatory constraints. \ac{FL} leverages this diversity without moving raw data, improving generalization while preserving confidentiality. Moreover, \ac{FedAvg} is communication efficient and robust under non-IID partitions, making it a practical baseline for clinical deployment \citep{Barbosa2025, Leonardo2025}.

\subsection{Image Preprocessing}\label{subsec:preproc}

We describe the input pipelines used before training and inference. The goals are: (i) harmonize files of varying size and color encoding so they are compatible with a pretrained ResNet-18 backbone, and (ii) in the preprocessed strategy, test whether standard intensity/denoising operations help or hinder performance in our federated setup.

\paragraph{Original (minimal processing)}
\begin{itemize}
  \item \textbf{Resize:} rescale each image to $224\times 224$.
  \item \textbf{Channels:} if single\textendash channel, replicate to three channels (for ImageNet initialization).
  \item \textbf{Normalization:} apply the same ImageNet mean and standard deviation normalization.
  \item \textbf{No intensity shaping:} no denoising, no histogram equalization, and no remapping beyond the normalization above.
\end{itemize}

\paragraph{Preprocessed (harmonization pipeline)}
\begin{itemize}
  \item \textbf{Grayscale standardization:} convert to single\textendash channel and replicate to three channels.
  \item \textbf{Resize:} rescale each image to $224\times 224$.
  \item \textbf{Pixel normalization:} scale raw intensities to $[0,1]$.
  \item \textbf{Light denoising:} apply a mild noise-reduction filter (e.g., small-kernel median or Gaussian).
  \item \textbf{Histogram equalization:} global equalization to broaden dynamic range in low-contrast slices.
  \item \textbf{Normalization:} apply the same ImageNet mean and standard deviation normalization.
\end{itemize}

\subsection{Test Time Augmentation}

Test Time Augmentation (\ac{TTA}) consists of applying data augmentation techniques during the inference phase rather than exclusively during training \citep{Mari2025}. Let $x$ denote an input image and $\mathcal{T} = \{T_{1}, T_{2}, \dots, T_{K}\}$ the set of stochastic transformations applied to $x$ (e.g., rotations, flips, or intensity adjustments). For each transformation $T_{k}$, the model $f_{\theta}$ produces a prediction as expressed in Equation~\eqref{eq:tta_pred}:  

\begin{equation}
\hat{y}_{k} = f_{\theta}(T_{k}(x)), \quad k = 1, \dots, K.
\label{eq:tta_pred}
\end{equation}

The final prediction $\hat{y}$ is then obtained by aggregating the $K$ outputs. For probabilistic outputs, the arithmetic mean is typically employed, as shown in Equation~\eqref{eq:tta_mean}:  

\begin{equation}
\hat{y} = \frac{1}{K} \sum_{k=1}^{K} \hat{y}_{k},
\label{eq:tta_mean}
\end{equation}
whereas for categorical predictions, majority voting is often used, as in Eq.~\eqref{eq:tta_vote}:  

\begin{equation}
\hat{y} = \arg\max_{c} \sum_{k=1}^{K} 1\{\hat{y}_{k} = c\},
\label{eq:tta_vote}
\end{equation}
where $c$ indexes the possible classes and $1\{\cdot\}$ is the indicator function.  

Prior work has grounded \ac{TTA} both theoretically and empirically in medical imaging. \citep{Wang2019} formalized \ac{TTA} for tumor segmentation, showing gains over conventional inference and providing more reliable uncertainty estimates. \citep{Islam2024} further positioned \ac{TTA} as strategy of modern augmentation pipelines, highlighting improvements in generalization. Building on these insights, we deploy \ac{TTA} in a federated classification setting to probe its effect on the accuracy and stability under client heterogeneity. Our use is model-agnostic and retraining-free: multiple stochastic views are aggregated at inference, and we compare the outcomes for models trained on original versus preprocessed MRI inputs.

\begin{figure}[ht!]
    \centering
    \resizebox{0.9\columnwidth}{!}{
    \begin{tabular}{cc}
        \includegraphics[width=0.45\linewidth]{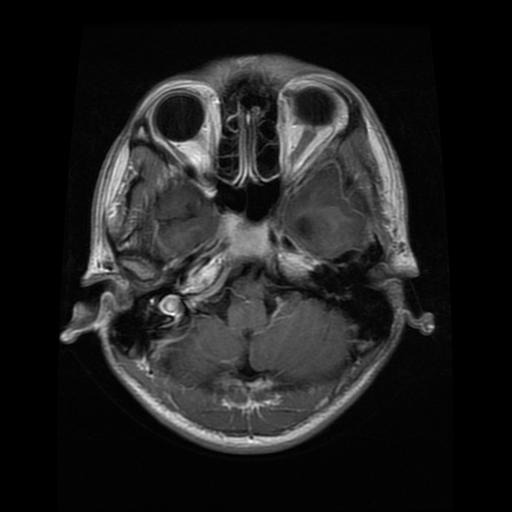} &
        \includegraphics[width=0.45\linewidth]{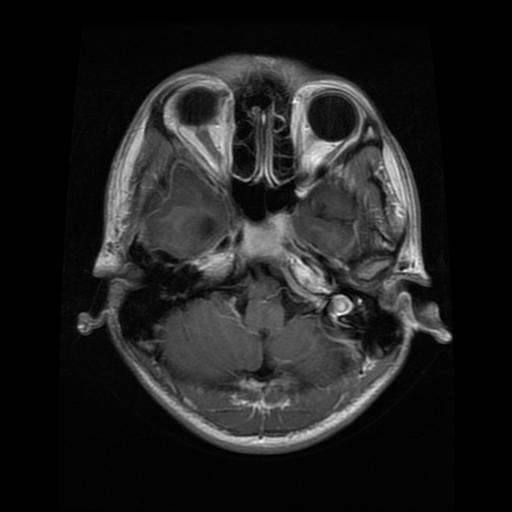} \\
        (a) Original & (b) Flip \\
        \includegraphics[width=0.45\linewidth]{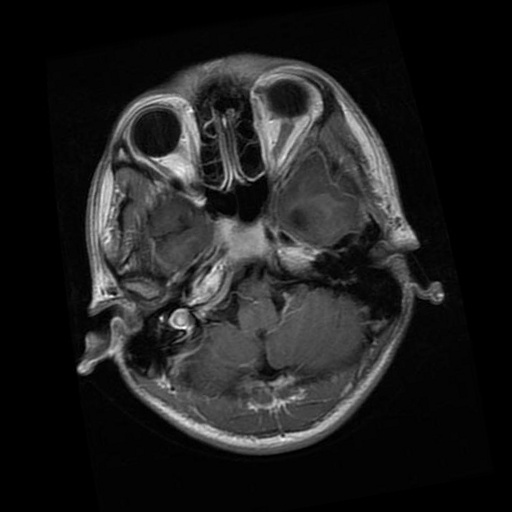} &
        \includegraphics[width=0.45\linewidth]{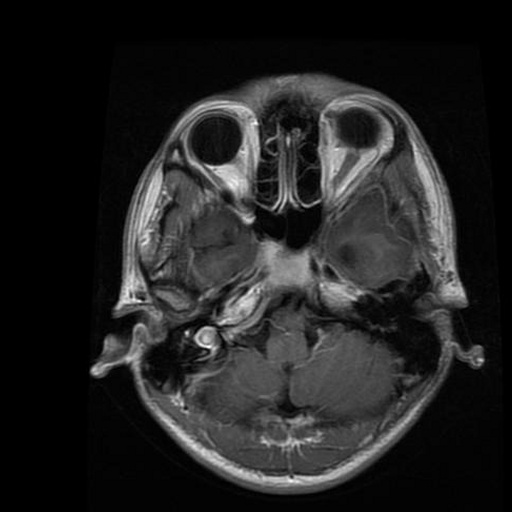} \\
        (c) Rotation & (d) Affine \\
    \end{tabular}
    }
    \caption{Original MRI scan and TTA-generated variations.}
    \label{fig:tta_examples}
\end{figure}

Figure~\ref{fig:tta_examples} illustrates the practical application of \ac{TTA} to an original \ac{MRI} sample. From a single input image, the process generates multiple stochastic views featuring transformations. These modifications maintain the anatomical integrity of the tumor region while introducing sufficient diversity to evaluate the model's robustness against realistic input variations. We fix the augmentation budget to \(K{=}10\) stochastic views per test image in our experiments.

\subsection{Evaluation Metrics}
We evaluate performance with standard classification metrics: Accuracy, Precision, Recall, and F1-Score \citep{Solanki2023}. Let $TP$ (true positives), $TN$ (true negatives), $FP$ (false positives), and $FN$ (false negatives) denote the entries of the confusion matrix.

\begin{itemize}
    \item \textbf{Accuracy:}
    \[
        Acc = \frac{TP + TN}{TP + TN + FP + FN}
    \]
    Measures the proportion of correctly classified instances relative to the total number of samples.

    \item \textbf{Precision:}
    \[
        Prec = \frac{TP}{TP + FP}
    \]
    Evaluates the proportion of positive predictions that are actually correct.

    \item \textbf{Recall:}
    \[
        Recall = \frac{TP}{TP + FN}
    \]
    Measures the model’s ability to correctly identify positive cases.

    \item \textbf{F1-Score:}
    \[
        F1 = 2 \cdot \frac{Precision \cdot Recall}{Precision + Recall}
    \]
    Represents the harmonic mean between Precision and Recall, particularly useful in scenarios with imbalanced classes.

\end{itemize}

\section{\uppercase{Results and Discussion}}\label{sec:results}

All experiments were performed on a workstation with an Intel(R) Core(TM) i5-8400 CPU @ 2.80\,GHz, 16\,GB RAM, and an NVIDIA GeForce RTX 2060 SUPER GPU. We used Python~3.8.10, PyTorch~2.4.1, and the Flower \ac{FL} framework~1.11.1~\citep{beutel2020flower}.

We first removed the exact duplicate images using the SHA-256 cryptographic hash. After deduplication, we created stratified train/test partitions (80/20) to preserve the class proportions observed in the full dataset in each split. For the federated setup, stratified subsets were assigned to each client, ensuring that (i) no image appeared in more than one split or client and (ii) both clients held representative samples from all classes. Because patient- or series-level metadata are unavailable, stratification was performed at the image level.

Both input strategies, Original and Preprocessed, operate on the same splits. In the Original strategy, images are consumed as provided (file decoding and tensor conversion only). In the Preprocessed strategy, we applied a harmonization pipeline prior to training.

For \(K>1\) augmented views, each test image is stochastically transformed using: resizing to \(224\times 224\), random horizontal flip (\(p=0.5\)), random rotation (\(\pm 10^\circ\), \(p=0.5\)), and a small random affine (translation \(\le 5\%\), scale \([0.95,1.05]\), \(p=0.5\)), followed by normalization. Predictions from the \(K\) views are aggregated by averaging the class probabilities or by majority voting. When \(K=1\), we apply only deterministic resizing and normalization. All preprocessing steps are deterministic within each strategy; the only stochasticity arises from \ac{TTA} at inference. Both federated clients used identically configured pipelines within their assigned strategies (Original vs.\ Preprocessed).

We trained for three communication rounds, with 50 local epochs per round and mini-batches of size 32. Optimization used Adam (learning rate \(0.001\)); aggregation used standard FedAvg. In each round, both clients participated (\(|\mathcal{C}_r|=2\); client fraction \(=1.0\)).
We evaluated both the Original and Preprocessed input strategies with and without \ac{TTA}, reporting descriptive metrics and paired statistical tests to assess the significance of performance differences across clients.

\subsection{Federated Baseline without TTA}\label{subsec:baseline-notta}

\begin{figure}[!ht]
  \centering
  \includegraphics[width=1\columnwidth]{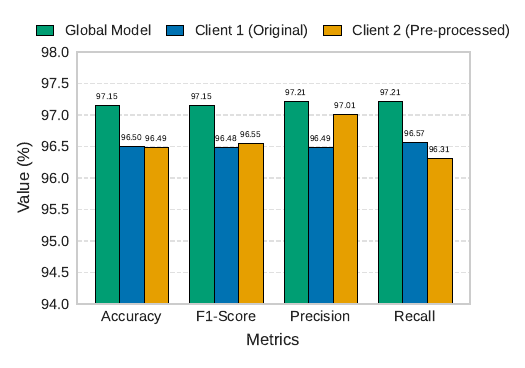}
  \caption{Performance metrics without TTA.}
  \label{fig:metrics_no_tta}
\end{figure}

We established a baseline by isolating the training-time processing without inference-time ensembling. The global model leads across all metrics, indicating that parameter aggregation via FedAvg yielded a stronger classifier than either client. Figure~\ref{fig:metrics_no_tta} shows the baseline performance of the global model and both clients without TTA and the global model surpasses either client across metrics.

The two clients produced comparable results; a small advantage appeared for the client trained on the original images (96.50\% accuracy) over the preprocessed client (96.49\%). 
Aggregation does more than smooth noise: it combines complementary error patterns to improve discrimination (Precision/F1) and coverage (recall). Effect sizes remain small, suggesting that per-client models are already strong under this dataset and partitioning, and that input processing choices primarily fine-tune calibration and operating points rather than driving large swings in accuracy.

\subsection{Effect of TTA}\label{subsec:effect-tta}

\begin{figure}[!ht]
    \centering
    \includegraphics[width=\columnwidth]{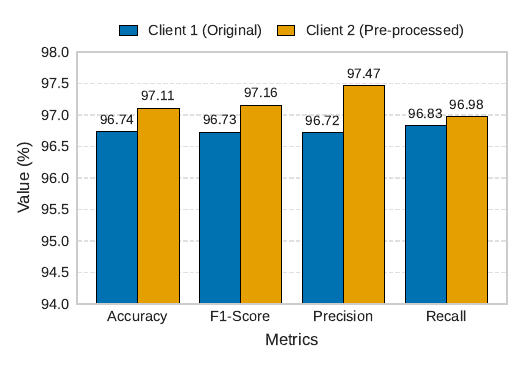}
    \caption{Performance comparison using TTA.}
    \label{fig:tta_metrics}
\end{figure}

The introduction of \ac{TTA} clarified the performance gap between the strategies. Figure~\ref{fig:tta_metrics} contrasts Client~1 (Original) and Client~2 (Preprocessed) with \ac{TTA}; the chart aligns with Table~\ref{tab:tta_results}.

With \ac{TTA}, the preprocessed client shows slightly superior performance (e.g., \(\approx 0.4-0.5\) percentage points), a modest but consistent margin that reflects gains in stability and precision from combining input standardization with augmented inference. This advantage likely stems from reduced noise and standardized inputs, which make predictions less sensitive to spatial and contrast variations. Concurrently, \ac{TTA} stabilizes the predictions and exposes small but systematic differences between the training configurations.

\begin{table}[!ht]
\centering
\caption{Results obtained with TTA (10 runs). Client 1 (original) and Client 2 (preprocessed).}
\label{tab:tta_results}
\renewcommand{\arraystretch}{1.3}
\resizebox{\columnwidth}{!}{%
\begin{tabular}{ccccc}
\hline
\textbf{C}                                                   & \textbf{Acc. (\%)}                                                   & \textbf{F1 (\%)}                                                     & \textbf{Prec. (\%)}                                                  & \textbf{Rec. (\%)}                                                 \\ \hline
\textbf{\begin{tabular}[c]{@{}l@{}} 1\end{tabular}}     & \begin{tabular}[c]{@{}c@{}}96.74 \\ $\pm$ 0.15\end{tabular}          & \begin{tabular}[c]{@{}c@{}}96.73 \\ $\pm$ 0.15\end{tabular}          & \begin{tabular}[c]{@{}c@{}}96.72 \\ $\pm$ 0.15\end{tabular}          & \begin{tabular}[c]{@{}c@{}}96.83 \\ $\pm$ 0.14\end{tabular}          \\ \midrule
\textbf{\begin{tabular}[c]{@{}l@{}} 2\end{tabular}} & \textbf{\begin{tabular}[c]{@{}c@{}}97.11 \\ $\pm$ 0.19\end{tabular}} & \textbf{\begin{tabular}[c]{@{}c@{}}97.16 \\ $\pm$ 0.18\end{tabular}} & \textbf{\begin{tabular}[c]{@{}c@{}}97.47\\  $\pm$ 0.16\end{tabular}} & \textbf{\begin{tabular}[c]{@{}c@{}}96.98 \\ $\pm$ 0.20\end{tabular}} \\ \hline
\end{tabular}%
}
\end{table}

\subsection{Statistical Analysis}

We evaluate whether preprocessing improves performance under a fixed test-time augmentation budget (TTA with \(K{=}10\)). Because each run yields paired outcomes for the two pipelines (Original vs.\ Preprocessed) under identical conditions, we test the mean paired difference with a two-sided paired \(t\)-test (\(n{=}10\) runs). Normality of per-run differences is checked with the Shapiro-Wilk test; if violated, we report the Wilcoxon signed-rank test. Effect size is reported as Cohen's \(d\) for paired designs \(\big(d=\bar{d}/s_d\big)\), where \(\bar{d}\) and \(s_d\) are the mean and standard deviation of the per-run differences. We test hypotheses as follows:
\bigskip
\begin{itemize}
  \item \textbf{Null hypothesis} (\(H_0\)): \\ No mean difference between conditions,
  \[
    H_0:\quad \mu^{(M)}_{\text{orig}} \;=\; \mu^{(M)}_{\text{pre}}.
  \]
  \item \textbf{Alternative hypothesis} (\(H_1\)): \\A mean difference exists between conditions,
  \[
    H_1:\quad \mu^{(M)}_{\text{orig}} \;\neq\; \mu^{(M)}_{\text{pre}}.
  \]
\end{itemize}
\bigskip

With \ac{TTA} fixed at \(K{=}10\), the preprocessed pipeline outperformed the Original pipeline by a modest yet consistent margin \((\Delta\text{Acc}=+0.45\ \text{pp})\).

Table~\ref{tab:tests_tta} indicates a significant accuracy gain for the preprocessed pipeline with TTA. The column ``Metric'' indicates the score compared (here, Accuracy); ``Diff.\ (pp)'' is the average difference in percentage points between preprocessed and original (positive favors Preprocessed); ``$t$ (df)'' is the paired $t$-test statistic with its degrees of freedom; ``$p$'' is the probability of observing such a difference under the null (smaller means stronger evidence); and ``Cohen’s $d$'' is the standardized effect size (larger means a more consistent gain).

\begin{table}[!ht]
\centering
\caption{Paired test at \(K{=}10\) (Original vs.\ Preprocessed).}
\label{tab:tests_tta}
\renewcommand{\arraystretch}{1.3}
\resizebox{\columnwidth}{!}{%
\begin{tabular}{@{}lcccc@{}}
\toprule
\textbf{Metric} & \textbf{Diff. (pp)} & \textbf{$t$ (df)} & \textbf{$p$} & \textbf{Cohen's $d$} \\
\midrule
Acc. & +0.45 & 5.72 (9) & $<$0.001 & 1.80 \\
\bottomrule
\end{tabular}
}
\end{table}

The results show that with TTA fixed at \(K{=}10\), the preprocessed pipeline is, on average, \(\,+0.45\) percentage points more accurate than the original pipeline. This improvement was consistent (Cohen’s \(d=1.80\), large effect) and very unlikely to be due to chance (\(p<0.001\)). 

With TTA, preprocessing helps and without TTA, the two pipelines perform essentially the same.
In the no-TTA condition, the two pipelines were statistically indistinguishable for Accuracy and F1 (\(p>0.05\)), reinforcing that preprocessing is beneficial primarily when paired with TTA.

\subsection{Addressing the Research Questions}

The experiments in Sections~\ref{subsec:baseline-notta} and \ref{subsec:effect-tta} answer our research questions. Table~\ref{tab:rq_summary} summarizes the findings. \\

\begin{table}[!ht]
\centering
\caption{Summary of findings.}
\label{tab:rq_summary}
\renewcommand{\arraystretch}{1.5}
\begin{tabularx}{\columnwidth}{@{}l X@{}}
\toprule
\textbf{RQ} & \textbf{Finding} \\
\midrule
\textbf{RQ1} & \textbf{Preprocessing helps only when paired with TTA.}
Without TTA, differences between the Original and Preprocessed pipelines were negligible; with TTA fixed at \(K{=}10\), the preprocessed pipeline showed a consistent, statistically significant gain (paired \(t\)-test, \(p<0.001\), \(d=1.80\)). \\
\addlinespace[0.5em]
\textbf{RQ2} & \textbf{TTA improves accuracy and stabilizes predictions.}
TTA increases performance and reduces variability across runs; it also reveals and amplifies the subtle advantage of preprocessed inputs, acting as a key stabilizer under federated inference. \\
\bottomrule
\end{tabularx}
\end{table}

\subsubsection*{RQ1: Does preprocessing help or hinder classification?}

Preprocessing helps classification only when TTA is applied. In the baseline federated setting without TTA (Section~\ref{subsec:baseline-notta}), the Original and Preprocessed pipelines performed comparably. With TTA at \(K{=}10\) (Section~\ref{subsec:effect-tta}), the Preprocessed pipeline significantly outperformed the Original pipeline (paired \(t\)-test, \(p<0.001\); see Table~\ref{tab:tests_tta}), indicating that the standardization benefits were realized at inference when augmented views were aggregated. \bigskip

\subsubsection*{RQ2: To what extent does TTA improve robustness and accuracy?}
TTA yields statistically significant improvements in accuracy and acts as a stabilization mechanism, reducing run-to-run variability and exposing systematic differences between training strategies (Section~\ref{subsec:effect-tta}). In practice, TTA should be the default inference strategy; when the computational budget permits, pairing TTA with light preprocessing delivers additional reliable gains.

\section{\uppercase{Conclusion}}\label{sec:conclusion}

This study assessed the effects of standard preprocessing and \ac{TTA} on federated \ac{MRI} tumor classification. Using two clients, Original and Preprocessed, we compared pipelines under a fixed augmentation budget and quantified paired mean differences. Preprocessing alone has a negligible effect, whereas preprocessing combined with \ac{TTA} yields consistent, statistically significant gains (e.g., +0.45 pp in accuracy; \(t(9)=5.72\), \(p<0.001\); Cohen’s \(d=1.80\)). In practice, \ac{TTA} should be the default inference strategy; when the budget permits, pairing it with light preprocessing delivers additional reliable improvements. 

In future work, we will scale to larger and more heterogeneous federations, adopt patient-level grouping when identifiers are available, and assess additional backbones (e.g., MobileNet, ResNet-34) to strengthen external validity. We will also broaden the augmentation policies and vary the number of \ac{TTA} samples to map accuracy-cost trade-offs under non-IID conditions. These steps are essential to guide reliable, privacy-preserving \ac{AI} for real-world clinical workflows.  

\section*{\uppercase{Acknowledgments}}
The authors gratefully acknowledges the financial support of FAPEMIG (Grant \#APQ00923-24). Andr\'e R. Backes gratefully acknowledges the financial support of CNPq (Grant \#302790/2024-1). 

\bibliographystyle{apalike}
\balance

{\small
\bibliography{refs}}



\end{document}